\documentclass[sigconf]{acmart}
\AtBeginDocument{%
  \providecommand\BibTeX{{%
    \normalfont B\kern-0.5em{\scshape i\kern-0.25em b}\kern-0.8em\TeX}}}

\copyrightyear{acmcopyright}
\acmYear{2023}
\acmDOI{}

\acmConference[KDD]{2nd Workshop on End-End Customer Journey Optimization at KDD 2023}{August 2023}{Long Beach, CA, USA}
\acmBooktitle{} 
\acmPrice{}
\acmISBN{}

\usepackage[ruled,vlined]{algorithm2e}
\usepackage{multirow}
\begin{document}

\title{B2B Customer Conversion Prediction: A Document Representation, Graph Theory, and CatBoost Driven Methodology}

\author{Tianqi Wang}
\authornote{Both authors contributed equally to this research.}
\orcid{}
\affiliation{%
  \institution{Purdue University}
  \streetaddress{610 Purdue Mall}
  \city{West Lafayette}
  \state{Indiana}
  \country{USA}
  \postcode{47906}}
  \email{}

\author{Sheikh Shams Azam}
\authornotemark[1]
\affiliation{%
  \institution{Purdue University}
  \streetaddress{610 Purdue Mall}
  \city{West Lafayette}
  \state{Indiana}
  \country{USA}
  \postcode{47906}}
  \email{}

\author{Wan Eih Huang}
\affiliation{%
  \institution{Purdue University}
  \streetaddress{610 Purdue Mall}
  \city{West Lafayette}
  \state{Indiana}
  \country{USA}
  \postcode{47906}}
  \email{}

\author{Anton Wiranata}
\affiliation{%
  \institution{HP Inc.}
  \city{Boise}
  \country{USA}}
  \postcode{}
  \email{anton.wiranata@hp.com}

\author{Christopher G. Brinton}
\affiliation{%
  \institution{Purdue University}
  \streetaddress{610 Purdue Mall}
  \city{West Lafayette}
  \state{Indiana}
  \country{USA}
  \postcode{47906}}
\email{cgb@purdue.edu}

\author{Jan P. Allebach}
\affiliation{%
  \institution{Purdue University}
  \streetaddress{610 Purdue Mall}
  \city{West Lafayette}
  \state{Indiana}
  \country{USA}
  \postcode{47906}}
  \email{allebach@purdue.edu}

\renewcommand{\shortauthors}{Wang and Azam, et al.}

\begin{abstract}
In the one-time selling B2B context, the buying cycle may last months or even years. During the long process, targeting customers that have a high potential to make purchases and recommending personalized campaigns accordingly are important for effective marketing. For this goal, we study the following problems, B2B customer data aggregation, customer feature generation, and prediction of whether a B2B customer would show interest in making a purchase (i.e., prediction of conversion into sales funnel). We propose an algorithm to aggregate individual contacts to the B2B customer level based on multiple keys. For non-standardized keys such as company names, we propose a novel architecture to cluster them in a domain encompassing irregularities such as spelling mistakes and spelling variants. We then define and generate a set of features and apply the CatBoost model for customer conversion prediction. Our framework achieves 91\% prediction accuracy. Based on the prediction results and analysis of the model, we then discuss personalized campaign recommendations to foster conversion.
\end{abstract}

\begin{comment}

\begin{CCSXML}
<ccs2012>
 <concept>
  <concept_id>10010520.10010553.10010562</concept_id>
  <concept_desc>Computer systems organization~Embedded systems</concept_desc>
  <concept_significance>500</concept_significance>
 </concept>
 <concept>
  <concept_id>10010520.10010575.10010755</concept_id>
  <concept_desc>Computer systems organization~Redundancy</concept_desc>
  <concept_significance>300</concept_significance>
 </concept>
 <concept>
  <concept_id>10010520.10010553.10010554</concept_id>
  <concept_desc>Computer systems organization~Robotics</concept_desc>
  <concept_significance>100</concept_significance>
 </concept>
 <concept>
  <concept_id>10003033.10003083.10003095</concept_id>
  <concept_desc>Networks~Network reliability</concept_desc>
  <concept_significance>100</concept_significance>
 </concept>
</ccs2012>
\end{CCSXML}

\ccsdesc[500]{Computer systems organization~Embedded systems}
\ccsdesc[300]{Computer systems organization~Redundancy}
\ccsdesc{Computer systems organization~Robotics}
\ccsdesc[100]{Networks~Network reliability}
\end{comment}
\keywords{Conversion Prediction, Word Clustering, Document Representation}


\maketitle

\section{Introduction}

Data analysis is a hot topic in B2B marketing. An increasing number of companies reported successful results from their big data and AI investments \cite{partners2021big}. Data analysis delivers value to firms by fostering B2B sales \cite{HALLIKAINEN202090} and strengthening business operations in supply chain \cite{GUNASEKARAN2017308} and customer relationship management (CRM) \cite{NAM2019233, ZERBINO2018818}. B2B marketing data involves large amounts of campaign records of individual contacts together with demographics and company characteristics (firmographics). Analyzing such data enables the possibilities to answer important business questions, such as, which customers are more likely to make purchases, are some sequences of campaign engagements are more effective than others in driving a sale. If so, is this sequence dynamic in nature, varying across the boundaries of regions and industries? 

The analytics of B2B marketing data, however, are facing difficulties in terms of data aggregation and customer feature selection. B2B marketing engagement data that shows the interactions between customer and company is usually recorded for each individual contact. To analyze customer behavior, it is important to collect all activities on the customer-company level, which includes the behavior of all involved employees. Even though CRM platforms can gather contacts for each customer account, missing engagement data exists, resulting from incomplete data. Typically, there are standardized keys and non-standardized keys linking contacts and customer companies. Standardized keys, like a Data Universal Number System (DUNS) number, are normally clean data, but it is difficult for all individual contacts and customer accounts to have such keys due to incomplete data or insufficient human labor. In order to have complete customer campaign engagement data and an unbiased dataset that covers various sizes of companies, regions, and industries, it is important to make use of non-standardized keys, such as company name and company location. In contrast to standardized keys, non-standardized keys can be free-entered text data which creates spelling mistakes, inconsistent abbreviations, and multi-language issues. Thus, the data processing pipeline should clean and standardize keys, and the mechanism of mapping contact to customer account should utilize multiple keys. 

LRFM (length, recency, frequency, and monetary) models \cite{anitha2019rfm, sarvari2016performance, 7057094, mesforoush2013customer} are popular for customer data analysis. Time series data is also widely used for customer analysis \cite{Espinoza2005, ibrahim2019}. However, when the B2B business is one or a few times of selling followed by supporting services, LRFM models are not applicable, and customer behavior time series data are not sufficient for modeling or analysis. In this case, we consider three types of customer attributes - demographics of employees from customer companies, firmographics, and customer campaign engagement features. Here, firmographic features refers to characteristics of organizations, companies, governmental entities, or any other type of firm. Demographic and firmographic features are extracted from databases, while campaign engagement data, even after aggregating contacts to the company level, are still individual records. We are interested in attributes such as the number of campaigns each customer has been engaged in and temporal campaign sequences. In a B2B context, it is not rare that multiple employees from one company participate in the same campaign. The feature generation mechanism should reflect both these facts, i.e., one same campaign and multiple employees. 

Considering the next steps when the customer with a high potential of conversion is targeted, although personalized content recommendation is actively studied in eCommerce, marketing campaign recommendations are rarely studied. Yang $et$ $al.$ \cite{yang1} proposed a graph-based next campaign to run (NCTR) recommender system for campaign recommendations. We discuss the actions that a B2B company can take to increase the customer conversion ratio based on our conversion prediction results.

Motivated by the needs of B2B business marketing data analysis, we address the problems of processing and aggregating individual contact data, prediction of conversion into sales funnel, and action recommendation based on conversion prediction results. Our study makes the following contributions. First, we propose a data preparation pipeline for data collection and cleaning. Our data preparation pipeline not only pre-processes data stored in the B2B databases but also collects online public firmographic data. To standardize keys, we design a novel architecture for text encompassing irregularities, including spelling mistakes and spelling variants. Second, we design an algorithm for aggregating contacts to the B2B customer level based on multiple keys. Customer companies having the same names may be different subsidiaries or branches and thus make their own decisions on purchases. Thus, we also consider location as a key. In order to utilize account ID, company name, and location as keys, we create two criteria and adopt connected components for contact aggregation. Third, we extract firmographic features and campaign engagement features and demonstrate that these features are effective for predicting conversion into the sales funnel. After feature extraction, we adopt the CatBoost model for conversion prediction. Last, we discuss the feasibility of recommending actions to foster sales based on conversion prediction results and model interpreting tools.

\begin{comment}

If only this type of key is considered to aggregate individual contacts, customer data will lose involved employees and campaign activities, especially for small companies and organizations that might not have a DUNS number.

For different tasks, these types of features are used in different ways. In B2C businesses, demographics and behavior features have been used in two separate phases for customer segmentation \cite{Namvar2010, Kandeil2014}. We use all types of features together in one phase for the task of conversion prediction. On the other hand, other tasks like campaign recommendation should distinguish different types of campaigns, for the reason that a B2B company can influence some customer actions through, for example, follow-up emails and recommending a next campaign, while it has less control over others, such as customer-initiated contact and document download. 

We then represent text documents in a vector space formed from the new vocabulary for further matching. Other techniques we adopt include translation API, fuzzy string matching, and connected components from graph theory. 

Text data like company names cannot be standardized by vocabulary checking because of the large number of made-up words. Our architecture tackles this issue by clustering words and using cluster indices as a new vocabulary.

We process and sort the aggregated company-level customer campaign activities into temporal sequences. Then we extract customer features from firmographic data and temporal campaign sequences for further analysis.
\end{comment}

\section{Related Work}

{\bfseries  Word clustering and cleaning} An extensive amount of work has been focused on word clustering and text clustering. Martin $et$ $al.$ \cite{MARTIN199819} employed an exchange algorithm using the criterion of perplexity improvement for bigram and trigram word clustering. Ushioda \cite{ushioda1996hierarchical} described a hierarchical clustering of words and clustering of multiword compounds. Word clustering in these works is developed for semantic analysis and based on context or word positions. In the task of company name text cleaning, there is no context available. Most spelling correction systems detect errors by searching in a dictionary or by semantic analysis of the surrounding context \cite{hladek2020survey}. A dictionary is also not available for the task of word clustering or text cleaning where made-up words are common.

{\bfseries Feature selection and extraction} Research by Kourou $et$ $al.$ \cite{KOUROU2015} and Guyon \& Elisseeff \cite{Guyon2003} emphasizes the importance of identifying the relevant attributes for predictive modeling. Studies on business analysis have been using financial data \cite{LOUZIS2012}, operational data \cite{GUNASEKARAN2004}, marketing data \cite{NGAI2009}, and textual data \cite{ZHAO2021} for various tasks. Venkatesh \& Anuradha \cite{Venkatesh2019} and Ghojogh $et$ $al.$ \cite{ghojogh2019feature} review general feature selection methods when a set of features are accessible. In our task, we gather collected data and web-scraped data, which contain customer behavior and company-level characteristics, and then apply feature selection and feature extraction techniques.

{\bfseries Conversion prediction} Conversion prediction has been studied for eCommerce \cite{moe2004dynamic, park2016investigating, yeo2017predicting}, and for B2C market in the domain of mobile advertising \cite{pereira2021multi, matos2019using}. These studies use customer online behavior or click-streams as input data, which is more accessible and complete. In the B2B market, few works are found for conversion prediction. Instead, researchers have explored B2B customer churn prediction \cite{figalist2019customer, gordini2017customers}. However, these studies are still mainly in the context of eCommerce.

{\bfseries Recommender system} Recommender systems have been an active topic in recent years. The major recommendation approaches include content-based methods, collaborative filtering, and hybrid methods \cite{adomavicius2005toward}. Temporal patterns are considered in the collaborative filtering model to learn the dynamic characteristics \cite{chen2013modeling}. In the B2B marketing campaign context, Yang $et$ $al.$ \cite{yang1} designed a social and temporal model for B2B marketing campaign recommendations. They exploited the temporal behavior patterns and constructed a temporal graph. The framework identifies common graph patterns and predicts missing edges in the temporal graphs. They also incorporated social factors into their approach.

\section{Dataset and Pre-processing} 

\begin{comment}
Campaigns in customer company data and campaigns in individual contact data are not repeated.

Many customer companies do not have extra campaign activities recorded in customer company data. 

Revenue and Employee Number can be either scalar values or value ranges. We use the median of a range as the value for Revenue or Employee Number if the web-scraped data is a range, and then convert all values to numbers. Web-scraped industries are of a large variety, while the industries in our dataset are sorted into seven segments. We adopt string processing, translation, and keyword mapping to standardize and sort all industries into these seven segments. 
\end{comment}

Some companies (including the one we are working with) have marketing data in two parts: individual contact data and customer company data. A summary of the marketing data we analyze is shown in Table \ref{tab:datasets}. The data is incomplete. In our dataset, some contacts are not linked to a customer company and thus have no Account ID. For both individual contact data and customer company data, firmographic features: Industry, Revenue, and Employee Number only exist in some contacts/companies.

\begin{table}
  \caption{Datasets}
  \label{tab:datasets}
  \small
  \begin{tabular}{p{0.25\columnwidth}p{0.34\columnwidth}p{0.34\columnwidth}} 
  % \begin{tabular}{cll}
    \toprule
    & Individual Contact & Customer Company\\
    \midrule
    \# of contacts/companies & 126.1 k & 11.4 k \\
    \# of campaign records & 148.4 k & 4.7 k \\
    Keys that exist for all contacts/companies & Company name, Location (Region, Sub-region, Country) & Company name, Account ID, Location (Region, Sub-region, Country) \\
    Keys that only exist for some of contacts/companies & Account ID, Location (City) & Location (City) \\
    Firmographics & Industry, Revenue, Employee Number (number of employees) & Industry, Revenue, Employee Number (number of employees)\\
    Campaign info & Campaign type, Campaign ID, Campaign time & Campaign type, Campaign ID, Campaign time \\
    Other info & Product the contact is interested in & Product the company is interested in \\
  \bottomrule
\end{tabular}
\end{table}

Our first objective is to aggregate contacts to companies such that we can gather all campaigns a customer participated in and then create temporal campaign sequences. However, the standardized key Account ID linking individual contacts to customer companies created by the marketing team only exists for some contacts. In order to complete the campaign data for customer companies that have Account IDs and create campaign data for customers that do not have Account IDs, we need other keys to determine whether individual contacts belong to one customer company. We use Company Name, Country, and City as extra keys to group contact data to the company level. These three types of data are all entered as free-form text responses. Country and City are cleaner data relative to Company Name. We apply data cleaning techniques to Country and City, and then the resulting data are directly used as keys. Our techniques and methods involve translating non-alphabetic Country and City to English through a translation API, converting to lowercase, removing special characters, removing customized stop-words such as ``city'', and manually checking. Company Name, on the other hand, has various made-up words and more irregularities such as spelling mistakes, inconsistent abbreviations, and multi-language issues. We describe the processing procedure for Company Name together with the proposed text document clustering architecture in the next section.

Learning from marketing experts' experience, firmographics are important attributes for marketing data analysis. Our datasets have missing values of these firmographic features in both contact data and company data. However, these data can be acquired in other databases or from online resources. We apply web scraping to extract firmographics in order to fill in this information. The web-scraped data and the firmographics in our dataset are not in the same formats. We adopt string processing, translation, and keyword mapping to standardize data from the two resources. After web scraping and pre-processing, there are still missing values in these firmographics. We set the global median in the processed data as Employee Number or Revenue for customer companies for which our dataset does not have this information. These firmographics will then be used in feature extraction and conversion prediction.

\section{Document Representation and Clustering Architecture}

\subsection{Architecture Overview}

\begin{figure*}[h]
  \centering
  \includegraphics[width=0.8\linewidth]{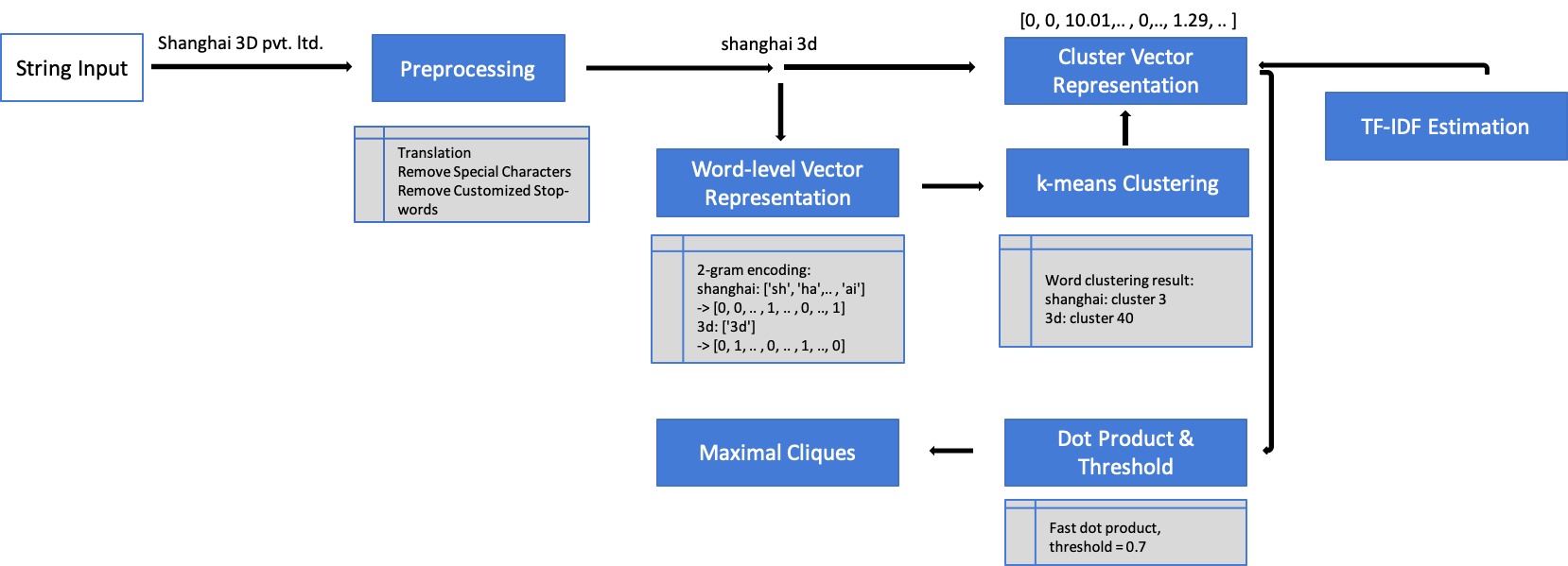}
  \caption{Architecture pipeline of our algorithm. All the constituent blocks are trained specifically to our data.}
  \label{Figure:document_clustering}
\end{figure*}

Text data like company names cannot be standardized by vocabulary checking nor represented directly by word embeddings because of the large number of made-up words. We tackle this issue by clustering words and using the cluster indices as a new vocabulary. Our architecture consists of a novel algorithm utilizing classical techniques such as stopword detection and correction, k-means clustering, and generation of vector representations in a vector space for the candidate documents. Figure \ref{Figure:document_clustering} summarizes the steps of the algorithm. It has two main phases: word-level clustering and string grouping in the cluster vector space.

\subsection{Word-level Clustering}

To standardize the representation of vocabulary words in text documents, we first conduct pre-processing and basic data-cleaning techniques on the text documents. Then we encode words and apply k-means clustering on the encodings. Using the clustering results, we then generate the standardized representation for vocabulary words using their cluster indices.

We pre-process the text documents, company names in our case, by the following steps: (1) convert text to lower case; (2) remove customized stop words, including \{the, company, co, pvt, ltd, llc, inc, corp, corporation\}, which do not provide information to distinguish the company names; (3) search and scrape the online public company profile using the processed company name from Step 2 as the search term. The web-scraping process for firmographics can be finished at the same time. This step utilizes a search engine and online public profile to remove some spelling variants and mistakes in the data; (4) translate non-English company names using the Translation API and compare scraped company names with search terms by their fuzzy ratio \cite{fuzzywuzzy2020}. If the ratio is larger than a threshold (we choose 70 in our case), use the scrapped company names for the next steps; (5) repeat Steps 1 and 2 on the scraped company names. Note that the search terms for automatic searching and web-scraping are before any language translation because small or medium-sized companies may not have profiles in multiple languages, and searching by the original languages can provide more accurate results.  

To cluster input strings (vocabulary words from company names in our case), we first aggregate all the input strings that are to be clustered and filter out repeated strings. These strings are then converted to their bigram representations, followed by the one-hot encoding of the bigram tokens. Bigram tokenization generated for the word “encoding” is the set $\{ \text{``en'', ``nc'', ``co''}, \cdots, \text{``in'', ``ng''}\}$. If the input strings only consist of 26 English letters and 10 Arabic numerals, then the one-hot encoded representation for each bigram has a dimension of $1,296$. Then each string, or vocabulary word, is represented by the sum of the one-hot encodings of all its bigram tokens. With the word-level encodings, we then perform Principal Component Analysis (PCA) and k-means clustering on principal components that retain 95\% variance. The cluster index is then the standardized representation of vocabulary words. We experiment with clustering with different values for $k$ and present the effect of $k$ in the word-level clustering in the next section.

\subsection{String Grouping in Cluster Vector Space}

After the word clustering, we infer the input strings using a bag of words wherein the words are the cluster indices from the previous word clustering results. We treat a cluster index as a vocabulary, then an input string is a list of cluster indices. We then vectorize the input strings by term frequency-inverse document frequency (TF-IDF) followed by normalization to lower the influence of common words such as city names. The dimension of the string vectors is the number of clusters of vocabulary words described in the previous section, denoted as $k$. After these steps, we have a vector representation of each input string.

Next, we evaluate the similarity of these vector representations using the dot product of the normalized vector representation. If the dot product between two vectors is larger than a threshold (we choose 0.7 in our case), the two strings are considered to be similar. Because the string vectors are sparse and of high dimensions, we adopt the fast dot product to accelerate the computation.

After fast dot product and thresholding, each string has 0, 1, or multiple similar strings. If we group all similar strings in one group, strings that are very different can end up in one group. For example, if String 1 is similar to String 2, String 2 is similar to String 3, ..., and String n-1 is similar to String n, then String 1 and String n can be very different but still be in one group. To overcome this issue, we cluster company names by maximal cliques of connected components. The steps are: (1) create a connected component graph where each node represents a company name and each edge represents the dot product of the two nodes larger than the threshold; (2) the largest maximal clique that has the most nodes is set as Group 1, and the strings corresponding to these nodes are regarded as the same; (3) remove these nodes in the maximal clique from the connected component graph; (4) repeat Steps 1, 2, and 3 and increase the group number until no node is left in the connected component graph. At the end of this step, strings that are in one group are considered to be the same.

\subsection{Experimental Results}

We first look at the effect of the number of clusters in the first phase through a set of examples. For the purpose of evaluation, we consider the values of $k \in \{10000, 20000, 30000, 40000, 50000\}$. It can be observed from Table \ref{table:word_clusters} that the clusters become less diverse as we increase the number of clusters. This leads to the trade-off between keeping $k$ too high and losing the variations in spelling or setting $k$ too low and allowing a lot of noisy clustering. For the following analysis, we consider $k=30000$.

\begin{table*}[h]
    \centering
    \caption{{Performance Comparison between Different Cluster Sizes.}}
    \small
    \begin{tabular}{ c c c c c}
      \toprule
      \textbf{$\boldsymbol{k}$-value} & \textbf{Example 1} & \textbf{Example 2} & \textbf{Example 3} & \textbf{Example 4} \\
      \midrule
       10000 & ['precision', 'bossprecision', 'precisionsteknik',& ['shanghai', 'shanghe', & ['argonne', 'argon', & ['machining',  \\
       & 'precisiion',  'precisionfab', 'precisionsignscom',  & 'shangcai', &  'falconcargo', & 'teaching', \\
       & 'osprecision', 'precicision', 'precisions', & 'shangai', &  'sargon', & 'phmachining', \\
       & 'precisione', 'precisionvetor', 'precisiondrone'] & 'shanghaibim'] & 'margoni'] & 'maching'],\\
      \midrule
      20000 & ['precision',& ['shanghai', 'shanghe', & ['argonne', & ['machining',  \\
       & 'precisiion',  'precisionfab',  & 'shangcai', & 'argon',  & 'phmachining', \\
       & 'precicision', 'precisions', & 'shangai', &  'sargon', &  'maching'], \\
       & 'precisione'] & 'shanghaibim'] & 'margoni'] & \\
      \midrule
      30000 & ['precision',& ['shanghai', & ['argonne', & ['machining',  \\
       & 'precisiion',  'osprecision',  & 'changhai'] & 'argon',  & 'machin', \\
       & 'precicision', & &  'sargon', &  'maching'], \\
       & 'precisions'] & & 'margoni'] & \\
       \midrule
      40000 & ['precision',& ['shanghai', & ['argonne'] & ['machining',  \\
       & 'precisiion',  'osprecision',  & 'changhai'] & & 'phmaching', \\
       & 'precicision', & & &  'maching'], \\
       & 'precisions'] & & & \\
       \midrule
      50000 & ['precision',& ['shanghai'] & ['argonne'] & ['machining',  \\
       & 'precisiion', & & & 'maching'] \\
       & 'precicision'] & & & \\
      \bottomrule
    \end{tabular}
    \label{table:word_clusters}
\end{table*}

\section{Contact Grouping Algorithm}

\begin{comment}
If we only consider Criterion 1, there are 6 customers for the 8 contact records. Contact 2 and Contact 3 are in the same company. If we only consider criterion 2, there are 5 customers. Contact 2 and Contact 3 are in different companies. However, we want any two contacts that satisfy either of the two criteria to be in one group. In that case, Records 1-6 are in one account, Record 7 is in one account, and Record 8 is in one account. 

A connected component is a subgraph in which any two nodes are connected to each other by paths or edges, and which is connected to no additional nodes in the rest of the graph. 
\end{comment}

Company Name is a common non-standardized key linking contacts and customer companies. However, Company Name itself is not sufficient to cluster individual contacts. One situation is that two subsidiaries of a large company in two different countries may be influenced by each other, but tend to make their own purchasing decisions. We set two criteria for contact grouping that make use of Company Name, Country, City, and Account ID: (1) contact records that have the same Company Name, the same Country, and the same City belong to one customer account; (2) contact records that have same Company Name, same Account ID, and same Country belong to one customer account. 

In the next section, we start by explaining how we apply a graph-based method and connected components, then we introduce our algorithm in pseudo-code.

\subsection{Connected Components Based Contact Grouping Algorithm}

\begin{figure}[h]
  \includegraphics[width=0.8\columnwidth]{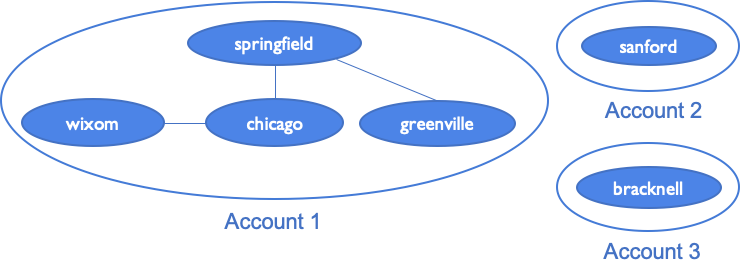}
  \caption{Contact grouping. Refer to Table 3 for contact data on which this grouping is based. The Springfield group is connected with the Chicago group because of common Account ID 001 and is connected with the Greenville group because of common Account ID 002. }
  \label{Figure: account}
\end{figure}

Table \ref{table: dummy data set} shows a dummy data set, which contains four fields we need for contact aggregation. Assume that we are grouping contacts that all have Company Name 3M. We propose a method to first group contacts by Criterion 1 and then gather the contact groups by Criterion 2. The first step can be easily achieved by conditioning Company Name, Country, and City. We apply the connected component in the second step. Figure \ref{Figure: account} shows how the contact groups are clustered to customer accounts in Step 2 by connected components. Each node is a group of contacts that have the same Company Name, the same Country, and the same City. In this graph, there are 3 connected components and therefore 3 customer accounts. 

\begin{table}[h]
%\footnotesize
\centering
\caption{A Dummy Data Set of Contacts.} 
\small
\begin{tabular}{p{0.07\columnwidth}p{0.18\columnwidth}p{0.24\columnwidth}p{0.14\columnwidth}p{0.14\columnwidth}} 
\hline
Index & Company Name & Country & City & Account ID  \\
\hline
  % [1ex] adds vertical space
  1 & 3M & united states & chicago & 001 \\
  2 & 3M & united states & wixom & - \\
  3 & 3M & united states & wixom & 001 \\
  4 & 3M & united states & greenville & 002 \\
  5 & 3M & united states & springfield & 001 \\
  6 & 3M & united states & springfield & 002 \\
  7 & 3M & united states & sanford & 003 \\ 
  8 & 3M & united kingdom & bracknell & 004 \\
\hline
\end{tabular}
\label{table: dummy data set} 
\end{table}

In our implementation, we perform the above-mentioned connected components-based method for every Company Name in every Country. Since both of the two criteria have Company Name and Country constraints, we iterate through Company Name and Country, and then we group contacts based on Account and City. There are 4 main steps in the algorithm: (1) create a mapping between City and contacts and another mapping between City and Account ID; (2) create the connected components graph where each node is a group of contacts that are in the same City, and each edge means the two contact groups share a common Account ID; (3) in every connected component, all related contacts are in one company; (4) for contact records in one company, we sort the campaign sequences in the temporal order. We also group records for which multiple people participate in one campaign and record the number of people from the company that participated in this campaign. We consider campaigns with the same Campaign ID to be one campaign. The details of the algorithm are summarized in the following pseudo-code (Algorithm 1) 

\begin{algorithm}[h]
\SetAlgoLined
\label{alg:connected_component}
\small
\For{every Company Name}{
  \For{every Country that the Company Name is related to}{
  Create a mapping between cities and contact records, denoted as Mapping 1\;
  Create a mapping between cities and Account IDs, denoted as Mapping 2\;
  Compare if there are common Account ID(s) between every two cities, using Mappings 1 and 2\;
  Create a graph where each node represents a city and each edge means a common Account ID exists between two connected cities\;
    \For{every connected component in this graph}{
    Find all related contact records by Mapping 1. These records are in one customer account\;
    Sort records by time, then by Campaign ID\;
    Create temporal campaign sequences and corresponding time sequences. In the meantime, group records for which multiple people attend one campaign.
    }
  }
}
 \caption{Connected Component Based Contact Grouping Algorithm.}
\end{algorithm}

\subsection{Experimental Results}

Our contact grouping algorithm applied the two criteria separately and resulted in 93,568 customer accounts. Table \ref{table: account roll-up results} shows an example of contact grouping results. Contacts in this customer account are from various cities, therefore City is left blank for this account.

\begin{table*}[!h]
\centering
\caption{Contact Grouping Result Example for a Single Contact Grouping.}
\label{table: account roll-up results}
\small
%\begin{tabular}{c c c c c c c} 
\begin{tabular}{p{0.23\columnwidth}p{0.15\columnwidth}p{0.15\columnwidth}p{0.3\columnwidth}p{0.27\columnwidth}p{0.27\columnwidth}p{0.22\columnwidth}} 
\hline
Company Name & County & City & Campaign Sequence & Time Sequence & Account ID & Contact Record Count \\
\hline
Company Name 1 & japan & &  [Email Campaign, Tradeshow, Tradeshow, Tradeshow, ...] & [FY2018-Q1, FY2018-Q2, FY2018-Q2, FY2018-Q3, ...] & [218814694.0, 218814694.0, 218814694.0, 218814694.0, ...] & 314 \\
\hline

\end{tabular}
\end{table*}

\begin{comment}
This algorithm applies to contact data when customer account-level temporal campaign sequences are created for the first time. After the account data is established, when a new contact campaign data touch is entered into the system, a method to add new data is needed. We look forward to studying this in the future.

\end{comment}

\section{Feature Extraction and Feature Selection}
\subsection{Feature Extraction}

After obtaining account data, we extract firmographic features and campaign activity-related features. Each contact record has its own firmographic information. Ideally, the process of contact grouping generates customer accounts having contacts with the same firmographic information for the same customer account. However, due to the incompleteness of the data and the fact that the contact grouping process is an approximation to the real case, an account can occasionally contain contacts that have firmographics that are different from those of other contacts in this account. We extract the most common values from all contact records in one account for firmographic features, while we retrieve campaign activity-related features as statistics either directly from activity sequences or by computation. For each customer account, we successfully extracted 156 features. Table \ref{table: Account Level Features} summarizes the descriptions of these features. We also sort 18 campaign types into 8 campaign categories and retrieve features from temporal campaign category sequences. The 4 main categories are digital campaigns like Webinar, in-person campaigns like Tradeshow, B2B seller-initiated connections like Email Campaign, and customer-initiated connections, such as Customer Connect with B2B Seller through Seller's website. 

These features can be used to train models to predict whether a customer can be converted into the sales funnel. The campaign activity sequences we have generated have all activities of people belonging to that customer account. However, not all these activities happened before conversion, for example, connecting the B2B seller through the seller's website for after-sales services. Therefore, we find conversion time and purchase time, and then truncate temporal campaign sequences for the task of conversion prediction.

\begin{comment}
For campaign-related features, we retrieve information or statistics from temporal campaign type sequences. 
\end{comment}

\begin{table*}[h]
\caption{Account Level Features.}
\label{table: Account Level Features}
\small
\begin{tabular}{p{0.4\columnwidth}p{0.31\columnwidth}p{1.24\columnwidth}} 
\hline
Feature & Extraction method & Description\\ 
\hline\hline
\emph{Firmographics} & & \\ 
Region$^1$ & Most common value & Location of a customer. There are in total 3 regions in our data. \\
Sub-region$^1$ & Most common value & Location of a customer. There are in total 18 sub-regions in our data. \\ 
Industry & Most common value & The industry a company is in. Customer company industries in our data are divided into 7 segments: industrial, mobility and transportation, consumer goods and electronics, healthcare, education and research, military, defense, and aerospace, and parts providers.\\  
Revenue & Most common value & Company revenue. If a revenue value is missing, the median value of all known revenue for companies in our data is used for this account.  \\ 
Employee number & Most common value & Total number of company employees in all sites. If an employee number is missing, the median value of all known employee numbers for companies in our data is used for this customer.  \\ 
Products & most common value & Product the customer is interested in. This is an estimation by the marketing team, not answered by the customer. \\
\hline\hline
\emph{Campaign related features} & &\\  
Campaign count & Statistic & Number of campaigns that people belonging to customer company have participated in. \\ 
People count & Statistic & Number of unique contacts in this company that have participated in campaigns.\\ 
Touch count & Statistic & Number of campaign touches, where one contact participating in one campaign counts as one touch.\\ 
Campaign count/Employee number & Calculation & Campaign count divided by Employee number \\ 
People/Campaign & Statistic & The average number of people from the customer company that participated in one campaign. We estimate this value by dividing the Touch count by the Campaign count. \\
Touch/People & Statistic & The quotient of Touch count by People count. We regard the result as the average number of campaigns a person in the account has participated in. \\ 
Campaign type count & Statistic & Number of campaign types of all campaigns people from a company have participated in. There are in total 18 campaign types in our data. \\ 
Ave \# of Qtr between successive campaigns & Statistic & Average number of quarters after a campaign until the next campaign. \\ 
Ave \# of Qtr before/after a campaign type/category & Statistic & The Average number of quarters before/after this campaign type/category after/until the previous campaign. \\ 
People/each campaign type & Statistic & Average number of people attending each type of campaign in the account. A value of 0 means the customer has not participated in that type of campaign. \\ 
Last/second to last/third to last campaign & Retrieve from campaign sequences & Last/second to last/third to last campaign in temporal campaign sequences. From marketing experts' insights, the most recent campaigns are important. \\ 
Campaign category sub-sequences & Retrieve from campaign category sequences & Number of sub-sequences of length 2 existing in the customers' campaign category sequences. We consider all possible length-2 sub-sequences and allow an arbitrary campaign category in between, e.g. sequences [B2B seller-initiated connection, digital] and [B2B seller-initiated connection, in-person, digital] both have a sub-sequence [B2B seller-initiated connection, digital].\\
\hline
\end{tabular}
\small
1: Location information is for a customer account, which can be different from the headquarters.
\end{table*}

\subsection{Feature Selection}

Feature selection is a process of reducing the number of input variables to reduce the computational cost and improve the performance of the model. In the 156 features we generated in the last step, there are 149 numerical features and 7 categorical features. We only conduct feature selection on the numerical features. The steps are as follows: (1) remove features that have one unique value over all accounts; (2) remove features that have a Pearson's correlation coefficient greater than 0.99 with other features; (3) train a tree-based model with all remaining features, record importance values, and remove features that have zero importance. The model we use is Light Gradient Boost Machine (LightGBM); (4) apply the sequential forward floating selection (SFFS) algorithm \cite{pudil1994floating} to select from the remaining features. 

We removed 23 features, each of which has only one unique value, 13 features to eliminate high correlation magnitude that is greater than 0.99 in our data, and 28 features with zero importance, which resulted in 85 numerical features remaining. SFFS then selected 33 features from these 85 features. Table \ref{table: selected features} lists the 33 features. Together with the 7 categorical features, Region, Sub-region, Industry, Products, last campaign, second to last campaign, and third to last campaign, we have 40 features in total. We examine the effect of feature selection on conversion prediction accuracy by training models on all 156 features and on 40 selected features, respectively, in the next section. 

\begin{table}[h!]
\caption{Selected Numerical Features.}
\small
\label{table: selected features}
\begin{tabular}{p{0.2\columnwidth}p{0.75\columnwidth}} 
\hline
Category & Numerical feature \\
\hline
Firmographics & Revenue  \\ 
Campaign statistics & Webinar count, Tradeshow count, Digital Campaign count, Email Campaign count, Conference Count, Open House count  \\ 
Campaign category sub-sequence & [digital, in-person],   [digital, customer-initiated connection],   [in-person, digital], [customer-initiated connection, digital] \\
Number of people & Touch count/Employees, People/campaign, Touch/people, People/Webinar, People/Tradeshow, People/Digital Campaign, People/Account Based Marketing (ABM), People/Social Media, People/B2B seller-initiated connection, People/digital, People/in-person  \\   
Time interval between campaigns & ave \# of Qtr between campaigns, \# of Qtr before Digital Campaign, \# of Qtr before ABM, \# of Qtr before Social Media, \# of Qtr after Content Syndication, \# of Qtr after Customer Connect With Seller by Website, \# of Qtr after Parts, \# of Qtr before B2B seller-initiated connection$^1$, \# of Qtr before digital$^1$ (category), \# of Qtr before in-person$^1$, \# of Qtr after in-person$^1$  \\
% \multirow{3}*{Firmographics} & Region \\
% & Subregion
% & Segment
\hline
\end{tabular}
\small 1: Features related to temporal campaign category sequences.
\end{table}

%%%%%%%%%%%%%%%%%%%%%%%%%%%%%%%%%%
% CatBoost
%%%%%%%%%%%%%%%%%%%%%%%%%%%%%%%%%%

\begin{comment}

 TS is an effective and efficient way to deal with  categorical features in which the category of feature i for k-th training example $x^i_k$ is substituted with one numeric feature equal to some target statistic $\hat{x}^i_k$. CatBoost choose ordered TS that satisfies the following two properties: (1) $\mathbb{E}(\hat{x}^i | y=v) = \mathbb{E}(\hat{x}^i_k | y_k=v) $ where ($x_k, y_k$) is the k-th training example. The expected feature values (TS) of feature i for a training example and a testing example should be equal when their target values are the same. (2) Effective usage of all training data for calculating TS features and for learning a model. Ordered boosting is proposed to tackle with the shift of gradient conditional distribution on a training example from that distribution on a test example, by applying similar ordering principle for TS. 
\end{comment}

\section{Conversion Prediction Based on CatBoost}

\subsection{CatBoost Model}

Prediction of which customers can be converted into the sales funnel enables marketing teams to take actions more effectively to convert customers. With the features we extracted in the previous section, one can build various machine-learning models to predict conversion. We choose the toolkit CatBoost \cite{prokhorenkova2019catboost} to build the model for the following reasons: (1) We have both categorical features and numerical features. CatBoost has an innovative algorithm for processing categorical features. (2) CatBoost is based on a decision tree, which makes the model easier to explain. There are two critical algorithmic advances introduced in CatBoost: ordered target statistics (TS) to encode categorical features, and ordered boosting based on gradient boosting algorithm with decision trees as base predictors (GBDT.

\subsection{Experiment and Results}

In our experiment, we evaluate CatBoost, random forest, logistic regression, and a recurrent neural network (RNN) on the prediction of customer conversion. In total, we have 93,568 customer accounts, of which only 15,724 have at least 2 campaign activities. Because one of our most important goals is to uncover how we can lead customers to conversion by influencing their activities, for this experiment we only use account data that has at least 2 campaign activities. Among these customer accounts, our data is unbalanced in terms of the target variable, i.e. conversion. For this reason, we use all data of 2,194 customers that are converted into the sales funnel and 2,194 randomly chosen customers from 13,530 unconverted accounts as our whole dataset. We randomly split the dataset into a training set (75\%) and a testing set (25\%) and use them for all these models.

The features used for these models and the results measured by accuracy are presented in Table \ref{table:comparison of models}. The parameters and metric for CatBoost are as follows: area under the curve (AUC) as the metric for overfitting detection and best model selection, maximum of 10 trees that can be built, 0.15 learning rate, and coefficient value 9 as the L2 regularization term of the cost function. Our RNN has one hidden layer of size 128 for each campaign activity step, and the output is produced through a fully-connected layer and log of the softmax function. This is a simple neural network structure. One possible future work is to improve the RNN structure for not only the task of conversion prediction but also the prediction of the next campaign activity. For logistic regression and random forest, we encode the categorical features with one-hot encoding. The CatBoost model trained on the 40 selected features reaches a conversion prediction accuracy of 91.0\%.

\begin{comment}
Random forest, logistic regression, and the RNN models have conversion prediction accuracies of 82.5\%, 75.2\%, and 77.4\%, respectively. Note that with parameter tuning, higher accuracy might be achieved by these models. 
\end{comment}

\begin{table*}[h!]
\caption{Comparison of Four Models.}
\label{table:comparison of models}
\small
\begin{tabular}{p{0.31\columnwidth}p{0.23\columnwidth}p{1.13\columnwidth}p{0.25\columnwidth}} 
\hline
& Feature \# & Features & Test Accuracy \\ 
\hline
Random Forest & \multirow{3}*{156 / 40}& \multirow{3}*{\begin{minipage}{1.14\columnwidth}We use the same features for the random forest, logistic regression, and CatBoost: 156 features before feature selection, 40 features after feature selection \end{minipage} } &  81.5\% / 82.5\%$^1$ \\
Logistic Regression && & 75.2\% / 74.7\%$^1$ \\ 
CatBoost & & & 90.2\% / 91.0\%$^1$ \\ 
RNN & 39 dynamic features + 9 static features & For the recurrent parts we use the one-hot encoding of the campaign type and job functions for each input step, of which the combined size is 39. Statistical features include Region, Employee number, Revenue, Segment/Industry, Products, Touch count, People count, Campaign count, and People/Campaign. & 77.4\% \\
\hline
\end{tabular}
\small 1: Accuracy on the left is for the model trained and tested on 156 features. The accuracy on the right is for the model trained and tested on 40 selected features.
\end{table*}

\begin{comment}
The authors showed that 3 simple properties uniquely determine a single solution of additive feature attributes: (1) Local accuracy. The explanation model matches the output of the original model for the simplified input. (2) Missingness. Features missing in the original input have no attributed impact. (3) Consistency. If a model changes so that some simplified input's contribution increases or stays the same, that input's attribution should not decrease. 
\end{comment}

\subsection{Action Recommendation Based on CatBoost Model and SHAP Values}

The CatBoost model is tree-based, which is more explainable than deep learning models. We adopt a model explaining tool, SHAP values, to explain the effects of features on conversion prediction. Since CatBoost has much higher accuracy than other models, we only apply the model explaining tool to the trained CatBoost model.

\begin{figure}[h]
%\centering
  \includegraphics[width=1\linewidth]{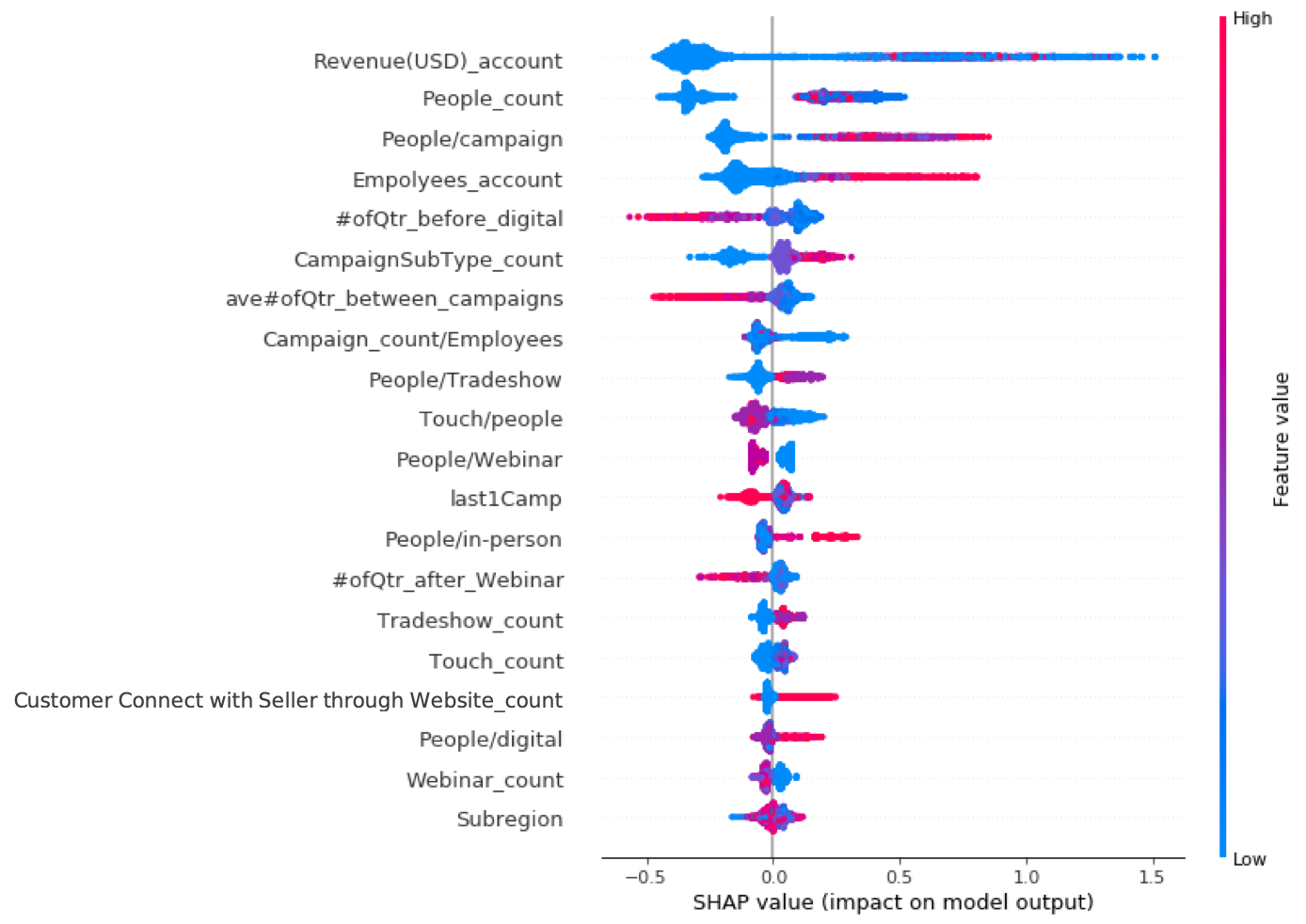}
  \caption{Summary Plot of SHAP Values. The horizontal location shows whether the effect of that value is associated with a higher or lower prediction. }
  \label{Figure: summary_shap}
\end{figure}

\begin{figure}[h]
\minipage{0.5\columnwidth}
  \includegraphics[width=\columnwidth]{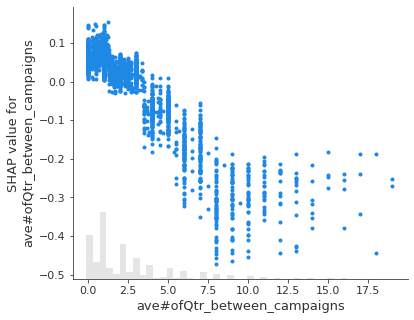}
\endminipage
\minipage{0.5\columnwidth}%
  \includegraphics[width=\columnwidth]{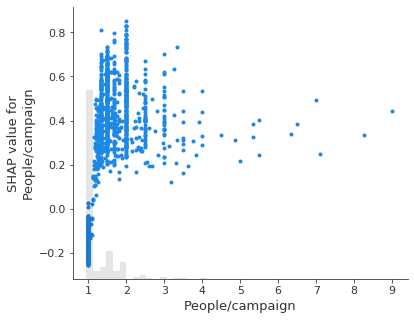}
\endminipage
\caption{SHAP values for the average number of quarters after a campaign until the next campaign, regardless of campaign types (left), and for the average number of people in an account attending each campaign (right). Vertical gray bars indicate the density of the data.}
\label{fig:shap_example}
\end{figure}

The SHAP values are proposed as a unified measure of the features' importance \cite{NIPS2017_8a20a862}. They comprise a class of additive feature importance measures, which have an explanatory model that is a linear function of binary variables. We choose TreeExplainer from the SHAP library, and fit this explainer to the trained CatBoost model. Then, we get SHAP values for the training data. Figure \ref{Figure: summary_shap} shows a summary of SHAP values for the top 20 important features. Features are ranked in descending importance order. Each dot represents a data point, and a dot cluster means a high data density for that feature (y-axis) around that SHAP value (x-axis). A higher feature value is shown with a magenta dot and a lower feature value with a blue dot. Positive SHAP values (to the right of the grey vertical line) are associated with higher prediction values (conversion), while negative SHAP values are associated with lower prediction values (no conversion). Figure \ref{fig:shap_example} shows SHAP values for two features. For every customer, an average time interval of less than or equal to 3 quarters after a campaign until the next campaign tends to have a positive SHAP value. Regardless of the campaign type, on average more than 1 person attending each campaign has a positive impact on conversion.

\begin{figure}[h]
\centering
  \includegraphics[width=1\linewidth]{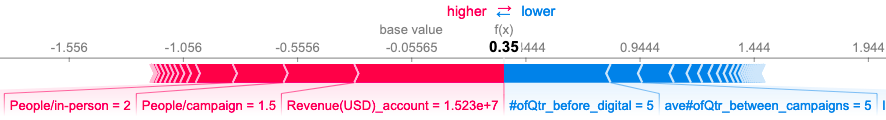}
  \caption{Visualization of Feature Impacts on One Account. The base value is the average of all output values of the model. 0.35 is the model output for this account.}
  \label{Figure: 000}
\end{figure}

When predicting whether a customer can be converted or not, we can apply the trained CatBoost model. If a customer has not been converted, actions can be taken to change some negative impacts. For example, a specific account has a negative prediction (unconverted), and the associated features' impact is shown in Figure \ref{Figure: 000}. Features in red positively influence the output, and features in blue negatively influence the output. We can encourage this customer to participate in campaigns within a shorter time interval instead of 5 quarters, as we know that SHAP values for less than or equal to 3 quarters after a campaign until the next campaign are positive. However, actionable recommendations developed from these findings are not guaranteed to make significant differences. One promising next step is to conduct causal inference to determine causality between variables.

\section{Conclusion}

In this paper, we present a framework for marketing data cleaning, aggregating contacts to customers at the account level, and predicting customer conversion into the sales funnel. We propose an algorithm to aggregate individual contacts to the B2B customer level based on multiple keys. For non-standardized keys such as company names, we propose a novel architecture to cluster them in a domain encompassing irregularities such as spelling mistakes and spelling variants. We demonstrate the effectiveness of predicting customer conversion by CatBoost with the features we developed. With CatBoost, we achieve a prediction accuracy of over 90\%. We also discuss the feasibility of recommending actions to foster sales based on conversion prediction results and model interpreting tools.


\begin{acks}
Research supported by HP, Inc. Boise, ID 83714.
\end{acks}

\bibliographystyle{ACM-Reference-Format}
\bibliography{references}

\appendix

\end{document}